\documentclass[11pt]{article}
\usepackage[hyperref]{ccl2024-en}
\usepackage{times}
\usepackage{url}
\usepackage{latexsym}
\usepackage{fancyhdr}

\usepackage{amsmath}
\usepackage{amsfonts}

\usepackage{algorithm}
\usepackage{algorithmic}

\usepackage{graphicx}
\graphicspath{ {./pics/} }
\usepackage{subfigure}

\usepackage{multirow}
\usepackage{booktabs}
\usepackage{makecell}
\usepackage{stfloats}
\newtheorem{Lem}{Lemma}
\DeclareMathOperator*{\argmax}{\arg\!\max}

\title{Cost-efficient Crowdsourcing for Span-based Sequence Labeling:\\Worker Selection and Data Augmentation}

\renewcommand\footnotemark{}

\author{Yujie Wang\textsuperscript{1,2*}\thanks{\textsuperscript{*}Equal contribution.} \quad
{\bf Chao Huang\textsuperscript{3*}} \quad
{\bf Liner Yang\textsuperscript{2,4\dag}\thanks{\textsuperscript{\dag}Corresponding author: Liner Yang}} \quad
{\bf Zhixuan Fang\textsuperscript{5}} \quad \\
{\bf Yaping Huang\textsuperscript{1}} \quad
{\bf Yang Liu\textsuperscript{2,4}} \quad
{\bf Jingsi Yu\textsuperscript{2,4}} \quad
{\bf Erhong Yang\textsuperscript{2,4}} \\
\textsuperscript{1}School of Computer and Information Technology, Beijing Jiaotong University, China \\
\textsuperscript{2}National Language Resources Monitoring and Research Center for Print Media, \\
Beijing Language and Culture University, China \\
\textsuperscript{3}Department of Computer Science, The University of California, Davis, USA \\
\textsuperscript{4}School of Information Science, Beijing Language and Culture University, China \\
\textsuperscript{5}Institute for Interdisciplinary Information Sciences, Tsinghua University, China \\
{\tt lineryang@gmail.com}
}

\date{}

\begin{document}

    \maketitle

    \begin{abstract}
        This paper introduces a novel crowdsourcing worker selection algorithm,
        enhancing annotation quality and reducing costs.
        Unlike previous studies targeting simpler tasks, this study contends with the complexities of label interdependencies in sequence labeling.
        The proposed algorithm utilizes a Combinatorial Multi-Armed Bandit~(CMAB) approach for worker selection, and a cost-effective human feedback mechanism.
        The challenge of dealing with imbalanced and small-scale datasets, which hinders offline simulation of worker selection, is tackled using an innovative data augmentation method termed \textit{shifting}, \textit{expanding}, and \textit{shrinking}~(SES).
        Rigorous testing on CoNLL 2003 NER and Chinese OEI datasets showcased the algorithm's efficiency, with an increase in $\textrm{F}_1$ score up to 100.04\% of the expert-only baseline, alongside cost savings up to 65.97\%.
        The paper also encompasses a dataset-independent test emulating annotation evaluation through a Bernoulli distribution, which still led to an impressive 97.56\% $\textrm{F}_1$ score of the expert baseline and 59.88\% cost savings.
        Furthermore, our approach can be seamlessly integrated into Reinforcement Learning from Human Feedback~(RLHF) systems, offering a cost-effective solution for obtaining human feedback.
        All resources, including source code and datasets, are available to the broader research community at \url{https://github.com/blcuicall/nlp-crowdsourcing}.
    \end{abstract}

    \section{Introduction}
    \label{intro}

    Crowdsourcing, the practice of obtaining labeled data from a multitude of contributors~\cite{howe2006rise}, has emerged as a pivotal tool in data collection for deep learning models.
    It offers a cost-effective alternative to expert labeling, making it especially valuable in today's data-driven research landscape~\cite{nowak2010reliable}.
    While its application spans various domains, from image labeling to text classification~\cite{venanzi2014community}, this paper narrows its focus on span-based sequence labeling tasks, which assign categorical labels to individual words within a sentence~\cite{erdogan2010sequence}.
    Notable examples of such tasks include named entity recognition (NER) and opinion expression identification (OEI)~\cite{collobert2011natural}.

    \cclfootnote{
    %
    %
        \hspace{-0.65cm}  
        \textcopyright 2024 China National Conference on Computational Linguistics

        \noindent Published under Creative Commons Attribution 4.0 International License

        \noindent This work was supported by the MOE (Ministry of Education in China) Project of Humanities and
Social Sciences (No. 23YJCZH264) and the funds of Research Project of the National Language Commission (No. ZDA145-17).
    }

    \begin{figure}
        \centering
        \includegraphics[scale=0.7]{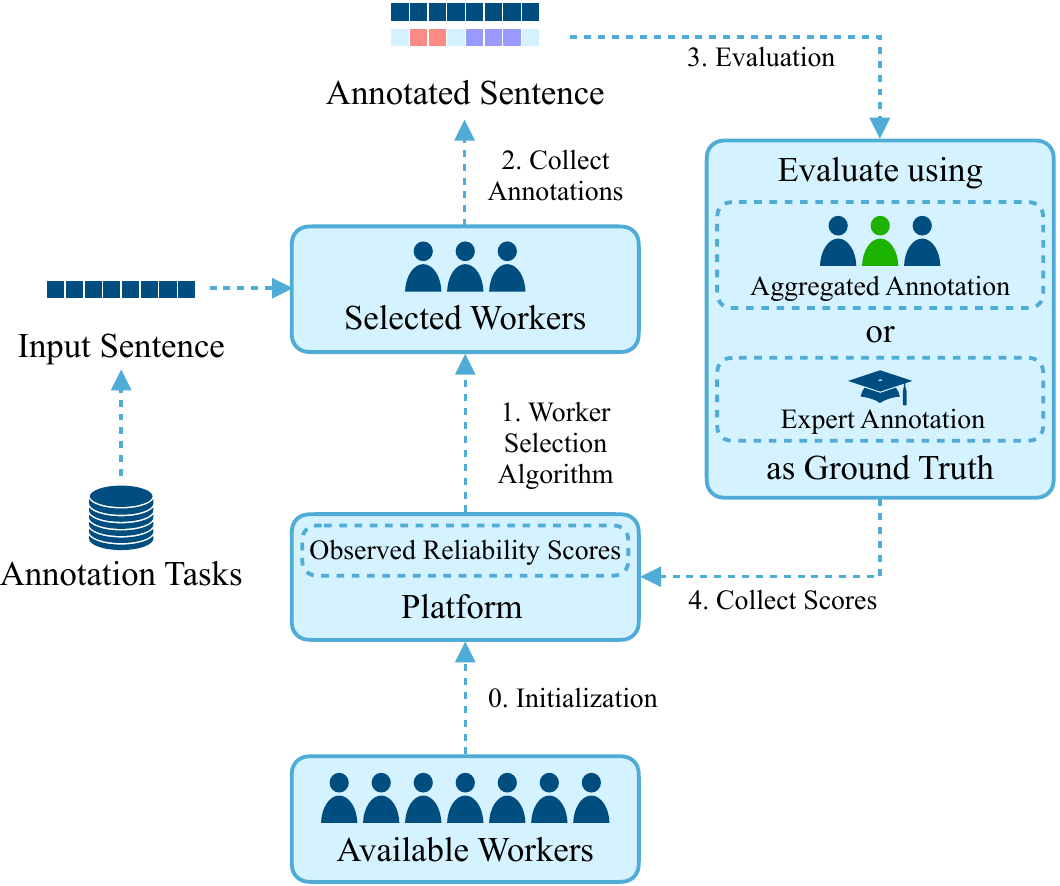}
        \caption{Our online worker selection framework for crowdsourcing.}
        \label{fig:framework}
    \end{figure}

    The inherent complexity of sequence labeling lies in the interdependencies of labels within a sequence.
    Unlike simpler tasks where labels are independent, sequence labeling requires contextual understanding, making it inherently more challenging~\cite{rodrigues2014sequence}.
    Consequently, annotations from crowd workers, who might not possess the expertise of trained annotators, often exhibit reduced accuracy.
    This underscores the imperative to enhance annotation quality, a challenge that this study addresses.

    A significant motivation driving this research is the potential application of a mixed feedback mechanism in Reinforcement Learning from Human Feedback (RLHF) systems.
    RLHF systems traditionally rely heavily on expert feedback, which, while accurate, is expensive and often not scalable~\cite{casper2023open}.
    By integrating feedback from both experts and aggregated crowd workers, we can achieve a balance between accuracy and cost.
    This hybrid approach not only maintains the quality of feedback but also significantly reduces the financial burden, making RLHF systems more accessible and scalable.

    Historically, research in this domain has concentrated on annotation aggregation~\cite{rodrigues2014sequence,nguyen-etal-2017-aggregating,simpson-gurevych-2019-bayesian}, employing methods post data collection.
    However, given the varied skill levels among crowd workers, a proactive approach that identifies and leverages the most accurate workers during the data collection phase can significantly enhance data quality.
    Termed as online worker selection, this strategy involves iterative allocation of a set budget across a pool of workers to optimize annotation quality~\cite{chen2013combinatorial}.
    This dynamic process grapples with the uncertainty of worker skill levels, necessitating a balance between exploring new workers and exploiting currently identified proficient ones.

    In the context of sequence labeling, traditional bandit-based algorithms~\cite{rangi2018multi} fall short due to the intricacies introduced by label dependencies.
    These intricacies manifest challenges in both annotation evaluation and aggregation.
    To address the evaluation challenge, this study employs the span-level $\textrm{F}_1$ score~\cite{derczynski-2016-complementarity}, a widely recognized metric, as the feedback signal in the worker selection process.
    The core \textbf{challenge} here is the accurate computation of the $\textrm{F}_1$ score in the absence of expert annotations as a reference.
    The objective is to minimize reliance on costly expert annotations.
    For aggregation, while the majority voting method is employed for its simplicity and effectiveness, its reliability can be compromised when faced with divergent annotations from different annotators.

    The overarching \textbf{goal} of this research is to maximize the quality of annotations while minimizing costs.
    This involves strategically replacing expert ground truth labels with aggregated crowd-sourced labels, ensuring that the overall $\textrm{F}_1$ score remains high.
    Such replacements are made only when there's a high level of agreement among crowd workers, indicating that expert evaluation might be redundant for that particular sequence. The proposed worker selection algorithm, as illustrated in Figure~\ref{fig:framework}, adopts an iterative approach: tasks are assigned to a subset of workers, their annotations are evaluated, and the resulting scores inform worker selection in subsequent rounds.

    However, real-world datasets present challenges due to their imbalanced nature and limited scale~\cite{rodrigues2014sequence,zhang-etal-2022-identifying}.
    Addressing these challenges, this paper introduces a data augmentation method tailored for span-based sequence labeling datasets.
    This method, designed to emulate potential human annotation errors, ensures that aggregated annotations remain meaningful.
    Three specific modifications, namely shifting, expanding, and shrinking, are applied to expert annotations, generating a spectrum of potential human annotations.
    This augmentation addresses dataset limitations, facilitating the offline evaluation of worker selection algorithms.

    In summary, this paper's contributions are manifold:
    \begin{itemize}
        \item It presents the exploration of worker selection for span-based sequence labeling tasks, recognizing the unique challenges they present.
        \item It employs the span-level $\textrm{F}_1$ score, evaluated by both experts and crowd workers, as a feedback mechanism, ensuring accurate worker selection.
        \item It introduces a data augmentation technique to counteract the limitations of real datasets, enabling effective offline simulations.
        \item Through rigorous experimentation, it demonstrates the efficacy of the proposed method, achieving impressive $\textrm{F}_1$ scores while significantly reducing expert annotation costs.
    \end{itemize}

    \setcounter{footnote}{0}

    \section{Related Work}\label{sec:related-work}

    Many studies~\cite{rodrigues2014sequence,rodrigues2018deep,nangia-etal-2021-ingredients} have used crowdsourcing for its efficiency and scalability.
    However, crowdsourcing suffers from the diversity of crowd workers' expertise and effort levels that are hardly measurable to task requesters.
    Different approaches to improving the quality of collected data have been proposed and studied.
    For span-based sequence labeling tasks, prior studies mainly focus on annotation aggregation.
    Rodrigues et al.~\shortcite{rodrigues2014sequence} proposed CRF-MA, a CRF-based model with an assumption that only one worker is correct for any label.
    HMM-crowd from Nguyen et al.~\shortcite{nguyen-etal-2017-aggregating} outperforms CRF-MA, but the effect of sequential dependencies is not taken into account.
    Simpson and Gurevych~\shortcite{simpson-gurevych-2019-bayesian} uses a fully Bayesian approach BSC which is proved to be more effective in handling noise in crowdsourced data.
    Aggregation methods are used \textit{after} the data collection process completes.
    But we aim to assure data quality and reduce cost \textit{during} collecting.
    To this end, we focus on worker selection in our paper.

    In online worker selection, we need to balance between exploring new workers and exploiting observed good workers.
    This exploration-exploitation tradeoff is extensively studied in the bandit literature~\cite{lai1985asymptotically}.
    In practice, we usually employ multiple crowd workers at the same time to finish the tasks more effectively.
    The combinatorial multi-armed bandit~(CMAB)~\cite{chen2013combinatorial} models this circumstance.
    Biswas et al.~\shortcite{biswas2015truthful} and Rangi et al.~\shortcite{rangi2018multi}
    reformulate the problem as a bounded knapsack problem~(BKP) and address it with the B-KUBE~\cite{tran2014efficient} algorithm.
    Song et al.~\shortcite{song2021minimizing} introduce empirical entropy as the metric in CMAB and minimize the cumulative entropy with upper confidence bound~(UCB) based algorithm.
    Li et al.~\shortcite{li2022harnessing} consider the scalability of worker selection on large-scale crowdsourcing systems.
    These studies propose different methods under the CMAB settings, but on more complex span-based sequence labeling tasks there exists no discussion.
    We present the study of worker selection with CMAB on span-based sequence labeling tasks and show that our work performs well on the quality and efficiency of data collection.

    \section{Methodology}

    Consider an online crowdsourcing system that can reach out to a group of crowd workers $W = \{w_1, w_2, \dots, w_N\}$.
    The workers are required to provide sequential annotations to a set of sentences $S = \{s_1, s_2, \dots, s_M\}$.
    More specifically, a worker annotates a sentence by assigning a tag from a finite possible tag set $C$~(e.g., a set of BIO tags~\cite{ramshaw-marcus-1995-text}) to each word.
    An annotation on sentence $s_i$ by worker $w_j$ is a tag sequence $\boldsymbol{a}_{ij} = a_1a_2\dots a_k\dots a_l$ where $a_k \in C$ and $l$ denotes the length of the sentence.
    We assume that every sentence is annotated by $K$ different workers independently.
    We define a task as the process of annotating one entire sentence, and hence there are in total $KM$ tasks.
    We seek to acquire an annotated dataset in which the average $\textrm{F}_1$ score of $\boldsymbol{a}_{ij}$ is maximized.
    If we know which workers give the best annotations a priori, we can simply ask these workers to finish all the tasks.
    However, such information is unavailable in practice, and we aim to design an algorithm that learns the best workers throughout the crowdsourcing process.

    In the beginning, we let each crowd worker annotate one sentence.
    We also ask the experts(e.g., well-trained linguists assumed to give the most precise annotations) to give one annotation for each of these sentences.
    Then we calculate the $\textrm{F}_1$ score of the annotation with the expert annotations as ground truth.
    We use these scores as the initial $\textrm{F}_1$ scores of workers.
    At each time step $t$ after initialization~(as illustrated in Figure~\ref{fig:framework}), we select a subset of workers $W_t \subset W$ to do annotation, based on criteria discussed in Section~\ref{sec:worker-selection-algorithm}.
    The size of the subset $W_t$ should be neither too big nor too small~(e.g., $0.3N$).
    We randomly choose a subset of sentences $S_t \subset S$, assign each $s_i \in S_t$ to $K$ different workers in $W_t$, and collect their annotations $\boldsymbol{A}_i = \{\boldsymbol{a}_{i1}, \boldsymbol{a}_{i2}, \dots, \boldsymbol{a}_{iK} \}, \forall i \in \{1, 2, \dots, \lvert S_t \rvert \}$.
    To evaluate workers' $\textrm{F}_1$ scores on $\boldsymbol{A}_i$, one can use the expert annotations as the ground truth, which, however, can be very expensive~\cite{iren2014cost}.
    To cut down this cost, we reduce the usage of expert evaluations whenever crowd annotations are similar enough.
    We use the Fleiss' Kappa score $\kappa$ to measure this similarity.
    The $\kappa$ score~($\kappa \leq 1$) is a statistical measure of inter-annotator agreement.
    A larger value of $\kappa$ indicates stronger agreement between the workers.
    $\kappa$ score exceeding an empirical threshold indicates that the crowd workers reach a consensus on $s_i$.
    In that case, we aggregate $\boldsymbol{A}_i$ with MV and use the aggregated annotation as the ground truth of sentence $s_i$.
    If the workers do not reach a consensus, we resort to expert annotations as ground truth.
    Next, we can calculate the $\textrm{F}_1$ scores of each $\boldsymbol{a}_{ij} \in \boldsymbol{A}_i$ and update the $\textrm{F}_1$ scores of the selected workers.

    \subsection{Problem Formulation}\label{sec:problem-formulation}

    At time $t$, we obtain $K$ crowd annotations $\boldsymbol{A}_i$ on each sentence $s_i \in S_t$.
    We denote all annotations collected on $S_t$ by $\boldsymbol{\mathcal{A}}_t = \{ \boldsymbol{A}_1, \boldsymbol{A}_2, \dots, \boldsymbol{A}_{\lvert S_t \rvert} \}$.
    To simplify our expression, we use $\textrm{F}_1^\textrm{Exp}(\boldsymbol{a}_{ij})$ to represent the $\textrm{F}_1$ score of $\boldsymbol{a}_{ij}$ using expert annotation as ground truth, and $\textrm{F}_1^\textrm{MV}(\boldsymbol{a}_{ij})$ to represent the $\textrm{F}_1$ score of $\boldsymbol{a}_{ij}$ using the MV aggregation of $\boldsymbol{A}_i \in \boldsymbol{\mathcal{A}}_t$ as ground truth.
    On collected annotation sets, $\textrm{F}_1^\textrm{Exp}(\boldsymbol{A}_i)$ denotes the average $\textrm{F}_1$ score of all $\boldsymbol{a}_{ij} \in \boldsymbol{A}_i$.
    Similarly, $\textrm{F}_1^\textrm{Exp}(\boldsymbol{\mathcal{A}}_t)$ denotes the average $\textrm{F}_1$ score of all $\boldsymbol{A}_i \in \boldsymbol{\mathcal{A}}_t$.
    As $\textrm{F}_1^\textrm{Exp}(\boldsymbol{\mathcal{A}}_t)$ reflects the true accuracy of crowd annotations, our objective is to maximize the average expectation, or equivalently the cumulative expectation of  $\textrm{F}_1^\textrm{Exp}(\boldsymbol{\mathcal{A}}_t)$ over time $T$.
    We formulate this problem as a CMAB problem below:
    \begin{align}
        \max \quad        & \sum_{t=1}^T \mathbb{E}[\textrm{F}_1^\textrm{Exp}(\boldsymbol{\mathcal{A}}_t)] \\
        \text{s.t.} \quad  & W_t \subset W, t \in \{1, 2, \dots, T\}
    \end{align}

    Since we have no information about workers' average $\textrm{F}_1$ scores, we need to balance exploring potentially better workers and exploiting the current best workers during worker selection.
    This tradeoff is extensively discussed in bandit literature where arms with unknown distributions form super-arms.
    The arms are associated with a set of random variables $X_{j,t}$ with bounded support on [0, 1].
    Variable $X_{j,t}$ indicates the random outcome of arm $j$ in time step $t$. The set of random variables $\{X_{j,t} | t \geq 1\}$ associated with arm $j$ are independent and identically distributed according to certain unknown distribution $D_j$ with unknown expectation $\bar{\mu}_j$.
    The platform plays a super-arm at each time step, and the reward of arms in it is revealed.
    These rewards are used as a metric for selecting the super-arm in future time steps.
    After enough time steps, the platform will be able to identify the best super-arm and keep playing it to maximize the overall reward.
    Similar to bandit terminologies, we call each worker $w_j \in W$ an arm and the worker subset $W_t \subset W$ a super-arm selected at $t$.

    \subsection{Worker Selection Algorithm}\label{sec:worker-selection-algorithm}

    Specifically, there are three methods to calculate the reward of worker $w_j$ at time step $t$ as follows.

    \paragraph{Expert Only}
    This is a benchmark approach where the $\textrm{F}_1$ score is calculated using only expert annotations as ground truth.
    This method provides intuitively the most accurate $\textrm{F}_1$ scores.
    The reward of worker $w_j$ is defined as:
    \begin{equation}
        \mu_j^\textrm{Exp}(t) = \textrm{F}_1^\textrm{Exp}(\boldsymbol{a}_{ij}(t))
    \end{equation}
    The expert-only method requires an expert annotation on every sentence, which is costly and usually not practical.

    \paragraph{Majority Voting (MV)}
    To reduce expert annotations, we aggregate $\boldsymbol{A}_i$ for each sentence $s_i$, and use the aggregated annotation via MV as ground truth, i.e.,
    \begin{equation}
        \mu_j^\textrm{MV}(t) = \textrm{F}_1^\textrm{MV}(\boldsymbol{a}_{ij}(t))
    \end{equation}

    \paragraph{Expert+MV}
    When workers give very different annotations on the same sentence (usually when the task is difficult), one can be uncertain about the voted (and possibly noisy) ground truth.
    In this case, we want to resort to both crowd workers and experts.
    The choice is based on the well-known Fleiss' Kappa score $\kappa$ that can quantitatively evaluate the agreement of crowd workers.
    For each sentence $s_i$, if $\kappa(\boldsymbol{A}_i)$ is greater than a preset empirical threshold value $\tau$, the reward of annotating workers is $\textrm{F}_1^\textrm{MV}(\boldsymbol{a}_{ij}(t))$.
    Otherwise, the reward is $\textrm{F}_1^\textrm{Exp}(\boldsymbol{a}_{ij}(t))$.
    In this way, MV is only used when the crowd workers can reach an agreement.
    Thus the reward is always calculated based on reliable ground truth.
    We summarize the reward of worker $w_j$ as:
    \begin{equation}
        \mu_j^\textrm{Exp+MV}(t) = \left\{
        \begin{aligned}
            \textrm{F}_1^\textrm{MV}(\boldsymbol{a}_{ij}(t)), \quad \kappa(\boldsymbol{A}_i) > \tau \\
            \textrm{F}_1^\textrm{Exp}(\boldsymbol{a}_{ij}(t)), \quad \kappa(\boldsymbol{A}_i) \leq \tau
        \end{aligned} \right.
    \end{equation}

    The $\epsilon$-Greedy, Thompson Sampling, and Combinatorial Upper Confidence Bound (CUCB) are three effective algorithms to solve the CMAB problem.
    For each worker $w_j \in W$, both algorithms maintain a variable $\bar{\mu}_j(t)$ as the average reward~(i.e., the average $\textrm{F}_1$ score) of worker $w_j$ at time step $t$.
    CUCB additionally maintains a variable $T_j(t)$ as the total number of sentences worker $w_j$ has annotated till time step $t$.
    Details of the worker selection algorithm with our \textbf{Exp.+MV} metric are shown in Algorithm~\ref{alg:cucb}.
    As for the selection criterion mentioned in the algorithm, $\epsilon$-Greedy utilize a hyper-parameter $\epsilon$ which refers to the probability of exploring random workers.
    Thus $1-\epsilon$ refers to the probability of exploiting the best workers till the current time step.
    Formally, $W_t$ is selected with a random variable $p \in [0, 1]$ as below:
    \begin{equation}
        \label{eqn:eg-selection}
        W_t = \left\{
        \begin{aligned}
            &\text{random}\ W_t \subset W, &\quad p < \epsilon \\
            &\argmax_{W_t \subset W} \sum_{w_j\in W_t}\bar{\mu}_j, &\quad p \geq \epsilon
        \end{aligned} \right.
    \end{equation}
    Thompson Sampling samples from gaussian distributions of workers' rewards at each time step $t$, and select workers which could maximize the total reward.
    CUCB handles the tradeoff by adding an item considering $T_j$ and $t$ to $\bar{\mu}_j$ like:
    \begin{equation}
        \label{eqn:cucb-selection}
        \begin{aligned}
            W_t &= \argmax_{W_t \subset W} \sum_{w_j\in W_t} \left(\bar{\mu}_j+\sqrt{\frac{3\ln{t}}{2T_j}}\,\right)
        \end{aligned}
    \end{equation}
    This makes workers with less annotations more likely to be selected as the algorithm proceeds.
    We provide a brief analysis in Appendix~\ref{sec:regret-analysis}.
    We explain on the application of our worker selection algorithms when building new datasets in Appendix~\ref{sec:new-dataset-explanation}

    \subsection{Data Augmentation Method}
    \label{sec:data-augmentation-method}

    \begin{figure*}[tp]
        \centering
        \includegraphics[scale=0.53]{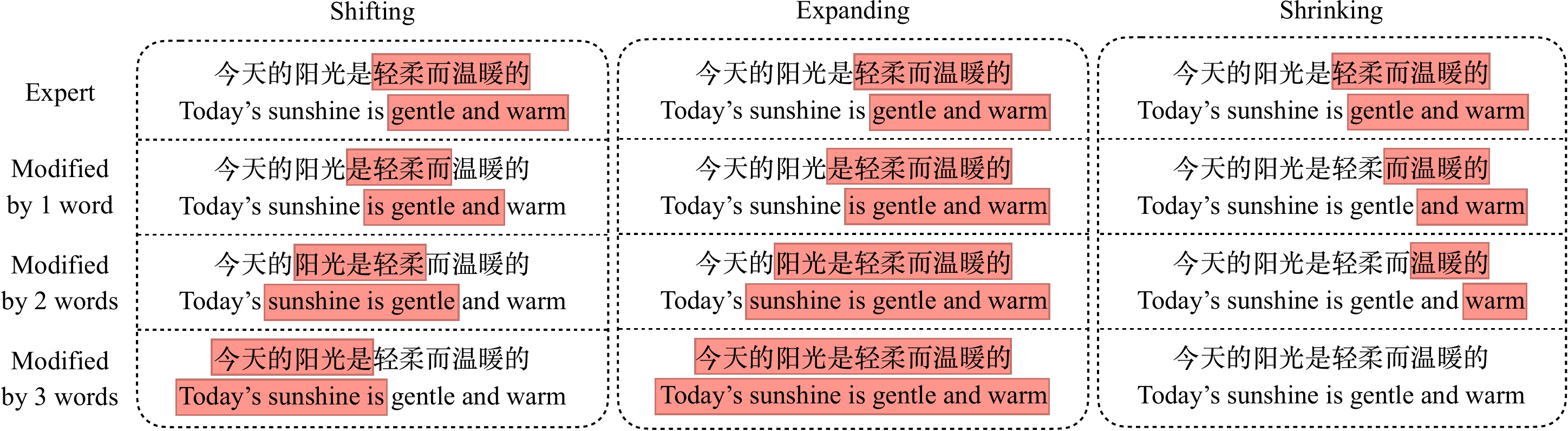}
        \caption{An example of the three methods to generate annotations. Chinese characters and corresponding English words with red backgrounds indicate annotation spans. }
        \label{fig:fakeannogen}
    \end{figure*}

    We propose the data augmentation method to facilitate the offline simulation of the crowdsourcing process, thus evaluating the worker selection algorithms.
    During offline simulation, when the worker selection strategy selects a certain worker to annotate a certain sentence, we can use the annotation in the original dataset if it exists.
    But if the selected worker did not annotate the sentence in the original dataset, we need to generate an annotation for the sentence.
    And the generated annotation should be in the same quality (depicted in F-score) as the real annotations by the worker.
    The generated annotation will be then used with the other annotations on the same sentence for majority voting.

    Generating the missing annotations for each worker $w_j$ is a great challenge when we expect the generated annotations to reflect the factual reliability of $w_j$.
    In other words, we expect the average $\textrm{F}_1$ score of each $w_j \in W$ to remain constant before and after augmenting the dataset with generated annotations.
    This is critical and difficult since real datasets are imbalanced and of small scale that cannot well support worker selection algorithms.

    As there lack previous work on generating missing crowd annotations for span-based sequence labeling, we start with several naive algorithms such as randomly generating label sequences as annotations, and mixing expert annotations with completely incorrect~(e.g., empty) annotations.
    But these algorithms either cannot produce annotations with expected $\textrm{F}_1$ scores, or generate confusing annotations which make later aggregation meaningless.
    This motivates us to design a data augmentation method specialized for span-based sequence labeling datasets.

    Through our statistical analysis and observation on the real datasets, we characterized the 3 most common annotation error patterns.
    Due to space limitation, we defer the detailed analysis to Appendix~\ref{sec:case-study}.
    Based on these analysis results, we propose a data augmentation method as follows:
    For each sentence $s_i \in S$, we modify the annotation span based on the expert annotation.
    We use three types of modifications to generate new annotation spans with different $\textrm{F}_1$ scores as illustrated in Figure~\ref{fig:fakeannogen}.
    The goal of these modifications is to simulate varying annotation errors made by human annotators.

    \paragraph{Shifting}
    We move both the left and the right border of the annotation span simultaneously in the same direction by one word per step.

    \paragraph{Expanding}
    We set one of the span borders fixed, and move the other border by one word per step to \textit{increase} the length of the annotation span.

    \paragraph{Shrinking}
    We set one of the span borders fixed, and move the other border by one word per step to \textit{decrease} the length of the annotation span.

    We perform these modifications on a span multiple times, generating new annotation spans, until (1)the modified span does not overlap with the original one, (2)one of the span borders reaches an end of sentence or another span in the same sentence, or (3) the span length becomes 0.

    For each sentence $s_i \in S$, $s_i$ may contain multiple annotation spans.
    We perform modifications on each span in $s_i$, and find all combinations of spans to form possible sentence annotations.
    With these methods, we can imitate crowd annotations with different kinds of errors in practice.
    Next, for each worker $w_j \in W_{ti}$, if $w_j$ has no annotation on $s_i$ in the original dataset, we select one from all the expert and generated annotations on $s_i$.

    We first calculate $\Bar{\varphi}_j$ as the average $\textrm{F}_1$ score of all annotations by $w_j$ on the original dataset, and then follow the detailed steps described in Algorithm~\ref{alg:fakeannopick} to do the selection.
    We aim to keep the overall $\textrm{F}_1$ score of $w_j$ unchanged.

    To better illustrate the procedure of the augmentation, we provide a running example in Appendix~\ref{sec:augmentation-running-example}.

    \section{Experiments}

    \subsection{Original Datasets}

    \begin{table}[tp]
    \centering
    \begin{minipage}{0.45\textwidth}
        \centering
        \begin{tabular}{cccccc}
            \toprule
            \textbf{Measure} & \textbf{Chinese OEI} & \textbf{CoNLL 2003} \\
            \midrule
            \# of Sentence   & 8047                 & 4580                \\
            \# of Worker     & 70                   & 47                  \\
            Span Length      & 5.05                 & 1.51                \\
            \midrule
            Max              & 658                  & 1626                \\
            Min              & 153                  & 48                  \\
            Range            & 505                  & 1578                \\
            Mean             & 368                  & 350                 \\
            Median           & 332.5                & 230                 \\
            SD               & 135.23               & 328.01              \\
            Variance         & 18286.52             & 107589.34           \\
            CV               & 36.71\%              & 93.57\%             \\
            \bottomrule
        \end{tabular}
        \caption{Statistics of the original datasets. Span lengths are averages. The terms SD and CV represent Standard Deviation and Coefficient of Variation respectively. The metrics Max, Min, Range, Mean, Median, SD, Variance, and CV pertain to the number of sentences annotated by each worker, indicating dataset imbalances.}
        \label{table:datasets}
    \end{minipage}
    \hfill
    \begin{minipage}{0.45\textwidth}
        \centering
        \begin{tabular}{c|p{1.55cm}<{\centering}p{1.55cm}<{\centering}p{1.55cm}<{\centering}}
            \toprule
            \bfseries\makecell{Worker \\ID}            & \bfseries\makecell{Rnd.  \\Gen.            \\$|\Delta\textrm{F}_1|$}            & \bfseries\makecell{SES  \\ Only            \\$|\Delta\textrm{F}_1|$}            & \bfseries\makecell{SES  \\+Alg.\ref{alg:fakeannopick}            \\$|\Delta\textrm{F}_1|$}            \\
            \midrule
            25 & 2.83 & 6.69 & 0.01 \\
            52 & 8.15 & 10.83 & 0.00 \\
            46 & 3.83 & 13.48 & 0.00 \\
            43 & 10.02 & 11.21 & 0.00 \\
            18 & 9.87 & 12.84 & 0.00 \\
            50 & 16.69 & 10.71 & 0.00 \\
            12 & 47.18 & 10.52 & 0.00 \\
            \midrule
            \textit{Avg.} & 14.08 & 10.90 & 0.0014 \\
            \bottomrule
        \end{tabular}
        \caption{Comparisons between different data augmentation methods on the error of span-level exact $\textrm{F}_1$ score of every crowd worker.
        The error $|\Delta\textrm{F}_1|$ is calculated as the absolute difference between each worker's $\textrm{F}_1$ score after augmentation and his real $\textrm{F}_1$ score.
        The methods \textbf{Rnd. Gen.}, \textbf{SES Only} and \textbf{SES + Alg.\ref{alg:fakeannopick}} are introduced in Section~\ref{sec:data-augmentation-method}.
        }
        \label{table:worker-scores-partial}
    \end{minipage}
\end{table}


    We compare our CMAB-based algorithms to several widely adopted baselines on two span-based sequence labeling datasets.

    \paragraph{CoNLL 2003}
    The CoNLL 2003 English named-entity recognition dataset~\cite{tjong-kim-sang-de-meulder-2003-introduction} is a collection of news article from Reuters Corpus~\cite{lewis2004rcv1}. The dataset contains only expert annotations for four named entity categories~(PER, LOC, ORG, MISC). Rodrigues et al.~\shortcite{rodrigues2014sequence} collected crowd annotations on 400 articles from the original dataset.

    \paragraph{Chinese OEI}
    The Chinese OEI dataset~\cite{zhang-etal-2022-identifying} consists of sentences on the topic of COVID-19 collected from Sina Weibo\footnote{\url{https://english.sina.com/weibo/}}, in which the task is to mark the spans of opinion expressions.
    The Chinese OEI dataset contains expert and crowd labels for two opinion expression categories~(POS, NEG).
    Detailed statistics are shown in Table~\ref{table:datasets}.

    \subsection{Data Augmentation}

    We augment both datasets with the method proposed in Section~\ref{sec:data-augmentation-method}.
    According to Table~\ref{table:datasets}, the most hard-working annotator in the OEI dataset provided annotations on 658 sentences, while the least one annotated only 153 sentences. On average, each crowd worker annotated 368 out of 8047 sentences in the Chinese OEI dataset. For the offline simulation of the worker selection process, we want every worker to annotate all 8047 sentences. Therefore we need to generate the missing 8047 - 368 = 7679 annotations for every worker, on average. This also applies similarly to the CoNLL 2003 dataset.

    \begin{figure}[tp!]
        \centering
        \subfigure[Chinese OEI]{
            \includegraphics[scale=0.45]{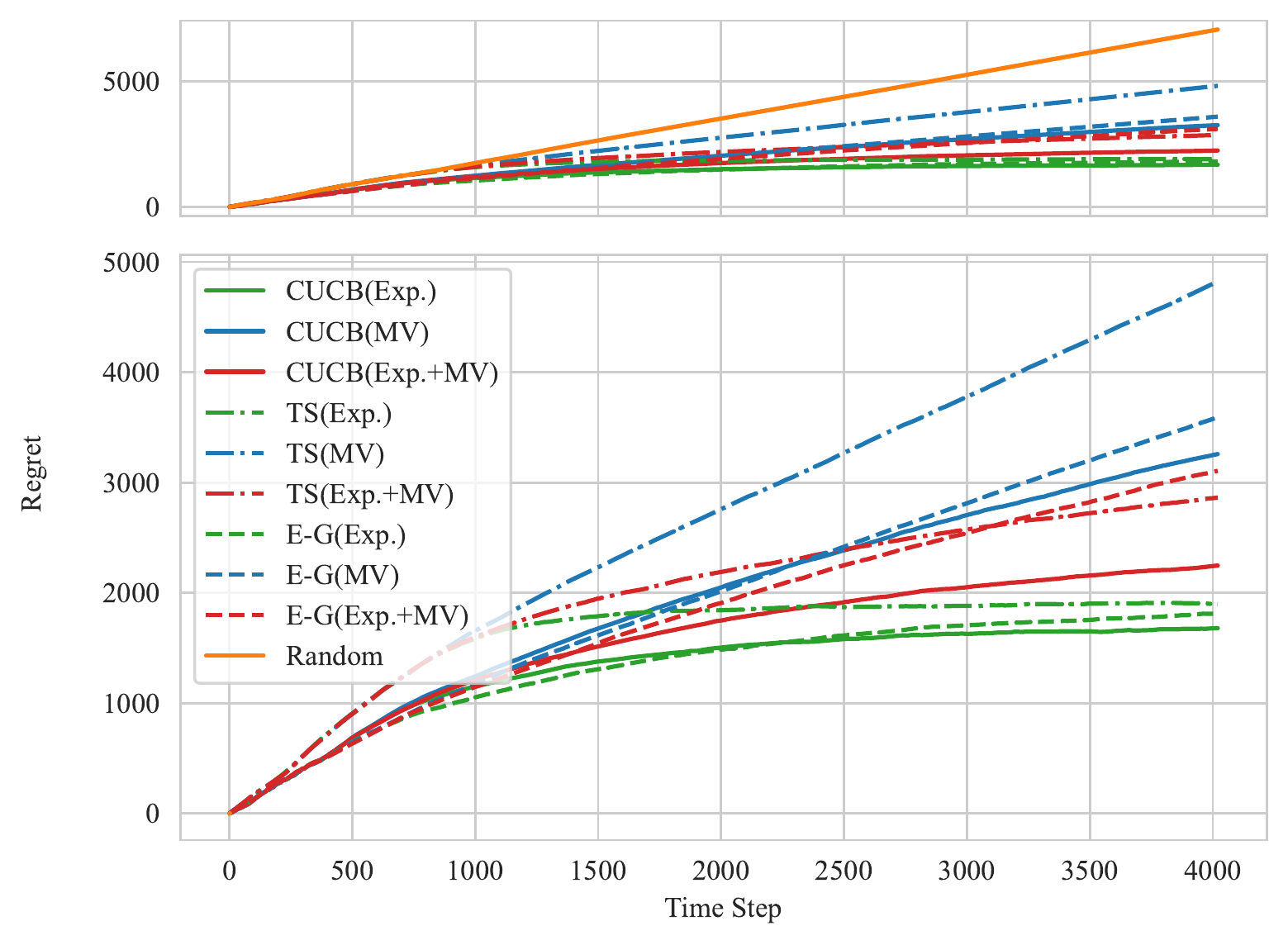}
            \label{fig:oei-regret}
        }
        \subfigure[CoNLL 2003]{
            \includegraphics[scale=0.45]{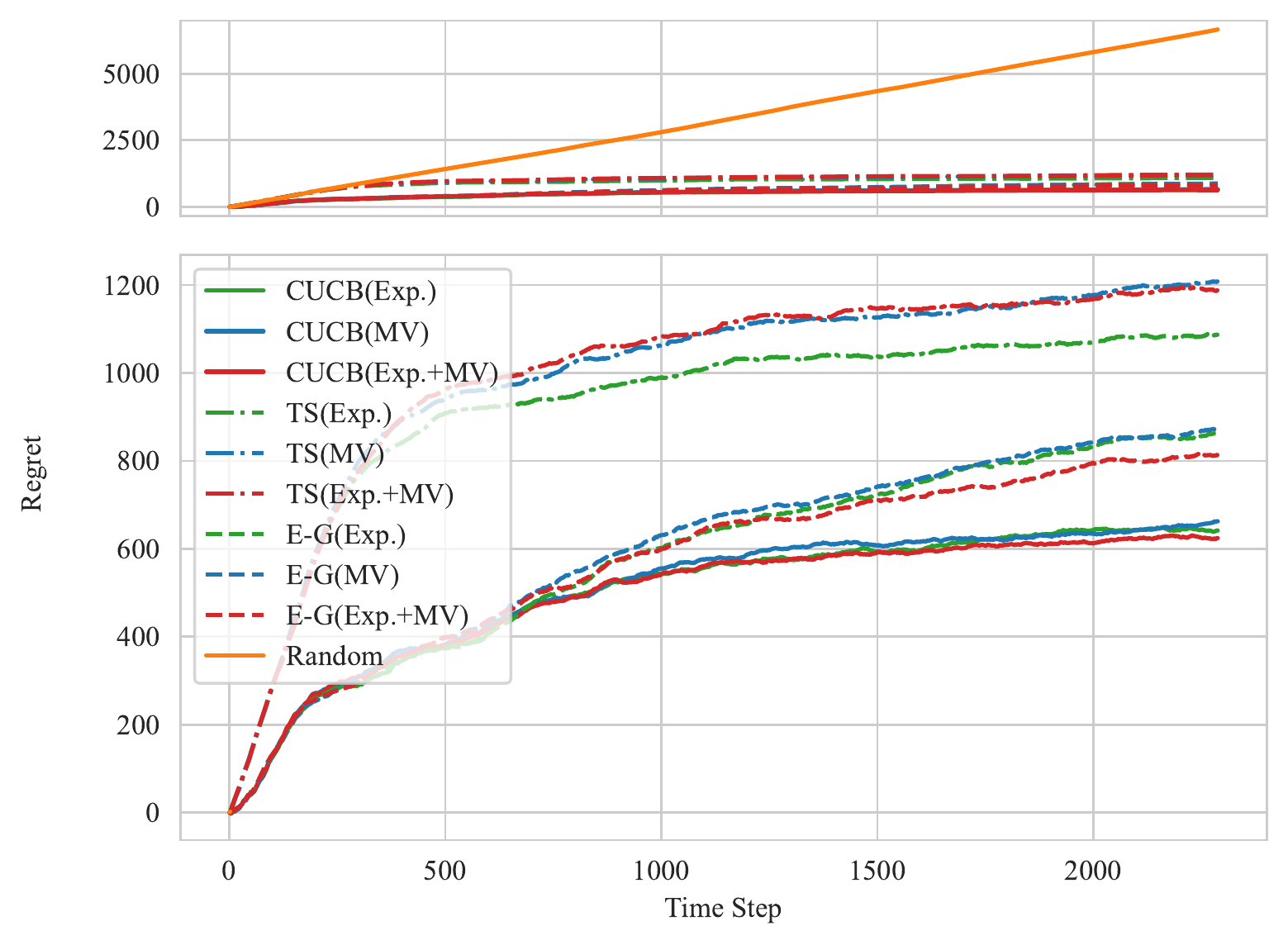}
            \label{fig:conll-regret}
        }
        \caption{Cumulative regrets w.r.t time steps of all different worker selection methods.}
        \label{fig:regrets}
    \end{figure}

    Through our method, the average $\textrm{F}_1$ score of each $w \in W$ remains nearly unchanged before and after augmenting the original dataset with generated annotations\footnote{The augmentation procedure takes about 2 hours on a computer with a 2.9 GHz Quad-Core Intel Core i7 CPU.}.
    Due to space limitation, we present comparisons of different augmentation algorithms with 10 sampled workers in Table~\ref{table:worker-scores-partial}.
    The complete results are deferred to Table~\ref{table:worker-scores} in the appendix.
    These results show that our \textbf{SES + Alg.\ref{alg:fakeannopick}} method clearly outperforms the other baselines, producing almost the same $\textrm{F}_1$ scores for each worker as their original ones.

    \subsection{Worker Selection}

    \paragraph{Baselines} We test the \textbf{Exp.+MV} method with 4 baselines: \textbf{Oracle}, \textbf{Random}, \textbf{Exp.}, and \textbf{MV}.
    \textbf{Oracle} always selects the empirical best super-arm $W^{opt}$ at every time step $t$.
    \textbf{Random} selects a different set of workers randomly at every time step $t$, which is equivalent to usual crowdsourcing procedure without worker selection.
    \textbf{Exp.}, \textbf{MV}, and \textbf{Exp.+MV} are CMAB-based algorithms introduced in Section~\ref{sec:worker-selection-algorithm}.
    The CMAB-based algorithms are tested with CUCB, Thompson Sampling and $\epsilon$-Greedy as the worker selection criterion respectively.

    \begin{table*}[tp!]
        \centering
        \begin{tabular}{l|ccc|ccc|ccc}
            \toprule
            \multirow{2}{*}{\textbf{Method}} & \multicolumn{3}{c}{\textbf{Token-level}} & \multicolumn{3}{c}{\textbf{Span-level Exact}} & \multicolumn{3}{c}{\textbf{Span-level Prop.}} \\
            \cmidrule{2-4}
            \cmidrule{5-7}
            \cmidrule{8-10}
            & $\textrm{P}$ & $\textrm{R}$ & $\textrm{F}_1$ & $\textrm{P}$ & $\textrm{R}$ & $\textrm{F}_1$ & $\textrm{P}$ & $\textrm{R}$ & $\textrm{F}_1$ \\
            \midrule
            Oracle                 & 65.69        & 83.99        & 70.00          & 78.15        & 72.23        & 74.96          & 87.97        & 80.03        & 83.82          \\
            Random                 & 55.95        & 66.42        & 57.50          & 64.42        & 55.64        & 59.40          & 75.70        & 62.61        & 68.54          \\
            \midrule
            $\epsilon$-G~(Exp.)    & 64.94        & 80.48        & \textbf{68.56} & 75.24        & 68.16        & \textbf{71.34} & 85.85        & 76.79 & \textbf{81.06} \\
            $\epsilon$-G~(MV)      & 64.44        & 80.22        & 67.98          & 74.69        & 67.59        & 70.77          & 85.67        & 76.09        & 80.59          \\
            $\epsilon$-G~(Exp.+MV) & 64.68        & 80.94        & \textbf{68.41} & 75.08        & 68.37        & \textbf{71.40} & 85.93 & 76.62 & \textbf{81.01} \\
            \midrule
            TS~(Exp.)              & 64.94        & 79.88        & \textbf{68.51} & 75.64        & 68.31        & \textbf{71.57} & 85.02        & 75.71        & \textbf{80.09} \\
            TS~(MV)                & 64.47        & 79.19        & 67.91          & 74.97        & 67.54        & 70.80          & 84.14        & 74.21        & 78.86          \\
            TS~(Exp.+MV)           & 64.20        & 79.09        & \textbf{67.62} & 75.27        & 67.83        & \textbf{71.12} & 84.77        & 75.39        & \textbf{79.81} \\
            \midrule
            CUCB~(Exp.)            & 65.65        & 80.34        & \textbf{69.24} & 75.94        & 69.12        & \textbf{72.20} & 86.17        & 77.22        & \textbf{81.45} \\
            CUCB~(MV)              & 65.39        & 80.00        & 68.91          & 75.95        & 68.90        & 72.08          & 86.13        & 76.67        & 81.12          \\
            CUCB~(Exp.+MV)         & 65.33        & 81.12        & \textbf{69.11} & 75.70        & 69.30        & \textbf{72.21} & 86.17        & 77.28 & \textbf{81.48} \\
            \bottomrule
        \end{tabular}
        \caption{Detailed $\textrm{P}$, $\textrm{R}$, and $\textrm{F}_1$ scores of all methods on the CoNLL 2003 dataset. All our algorithms perform significantly better than the Random (i.e., naive crowdsourcing) baseline.}
        \label{table:conll-results}
    \end{table*}

    \begin{table*}[tp]
        \centering
        \begin{tabular}{l|ccc|ccc|ccc}
            \toprule
            \multirow{2}{*}{\textbf{Method}} & \multicolumn{3}{c}{\textbf{Token-level}} & \multicolumn{3}{c}{\textbf{Span-level Exact}} & \multicolumn{3}{c}{\textbf{Span-level Prop.}} \\
            \cmidrule{2-4}
            \cmidrule{5-7}
            \cmidrule{8-10}
            & $\textrm{P}$ & $\textrm{R}$ & $\textrm{F}_1$ & $\textrm{P}$ & $\textrm{R}$ & $\textrm{F}_1$ & $\textrm{P}$ & $\textrm{R}$ & $\textrm{F}_1$ \\
            \midrule
            Oracle                 & 62.88        & 68.62        & 64.80          & 54.48        & 51.97        & 53.07          & 72.79        & 64.07        & 68.15          \\
            Random                 & 58.49        & 57.30        & 57.42          & 43.99        & 35.50        & 39.18          & 69.01        & 52.36        & 59.55          \\
            \midrule
            $\epsilon$-G~(Exp.)    & 61.91        & 64.58        & \textbf{62.61} & 51.72        & 46.37        & \textbf{48.76} & 72.28        & 60.25 & \textbf{65.72} \\
            $\epsilon$-G~(MV)      & 60.87        & 63.52        & 61.55          & 48.72        & 44.66        & 46.37          & 70.15        & 58.94        & 64.05          \\
            $\epsilon$-G~(Exp.+MV) & 61.76        & 64.46        & \textbf{62.47} & 49.14        & 45.35        & \textbf{46.96} & 71.21 & 59.92 & \textbf{65.08} \\
            \midrule
            TS~(Exp.)              & 62.66        & 64.91        & \textbf{63.20} & 49.76        & 42.34        & \textbf{45.69} & 72.15        & 60.20        & \textbf{65.63} \\
            TS~(MV)                & 59.82        & 61.90        & 60.25          & 44.81        & 40.71        & 42.36          & 67.72        & 56.05        & 61.34          \\
            TS~(Exp.+MV)           & 61.66        & 64.03        & \textbf{62.23} & 47.20        & 42.36        & \textbf{44.49} & 70.66        & 59.07        & \textbf{64.35} \\
            \midrule
            CUCB~(Exp.)            & 63.02        & 63.75        & \textbf{62.93} & 52.24        & 45.51        & \textbf{48.56} & 73.05        & 59.53        & \textbf{65.60} \\
            CUCB~(MV)              & 61.94        & 62.09        & 61.55          & 49.57        & 44.39        & 46.66          & 71.22        & 57.59        & 63.68          \\
            CUCB~(Exp.+MV)         & 62.83        & 63.62        & \textbf{62.75} & 51.31        & 45.60        & \textbf{48.16} & 72.48        & 59.33 & \textbf{65.25} \\
            \bottomrule
        \end{tabular}
        \caption{Detailed $\textrm{P}$, $\textrm{R}$, and $\textrm{F}_1$ scores of all methods on the Chinese OEI dataset. All our algorithms perform significantly better than the Random (i.e., naive crowdsourcing) baseline.}
        \label{table:oei-results}
    \end{table*}

    \paragraph{Regret as a Metric} We evaluate our worker selection algorithms using cumulative regret, a metric indicating the performance deviation from the oracle's selection defined as:
    \begin{equation}
        \label{eq:regret}
        R(T) = \sum_{t=1}^T \left( \sum_{w_j \in W^{opt}} \bar{\mu}_j - \sum_{w_k \in W_t} \mu_k(t) \right)
    \end{equation}
    In our experiments, we request 10 annotations per sentence, allowing CMAB-based algorithms to converge, and select 20 workers at each time step $t$.
    On the Chinese OEI dataset, setting the kappa threshold $\tau$ to 0.4 in \textbf{Exp.+MV} results in a 57.02\% reduction in expert annotation cost, while a 0.65 threshold on the CoNLL 2003 dataset leads to a 43.83\% cost reduction.

    Results show \textbf{Random} consistently underperforms across datasets.
    On the Chinese OEI dataset, \textbf{Exp.+MV} surpasses \textbf{MV}, albeit with higher regret than \textbf{Exp.}, justified by the substantial cost savings.
    On the CoNLL 2003 dataset, \textbf{Exp.+MV} even outperforms \textbf{Exp.}, suggesting crowd workers can provide valuable input for simpler tasks like NER.
    Overall, algorithms employing the CUCB criterion demonstrate superior performance, with \textbf{CUCB~(Exp.+MV)} excelling in balancing cumulative regret and expert cost.

    \paragraph{Effect of $\tau$ on $\textrm{F}_1$ and cost} Next, we discuss how different kappa threshold values $\tau$ affect the average $\textrm{F}_1$ score of the produced annotation dataset.
    We test $\tau \in [0, 1]$ with a step of 0.05.
    In real datasets like CoNLL 2003 and Chinese OEI, the number of annotations per sentence is often quite small.
    To better fit the practical situations, we ask for 4 annotations on each sentence in the following experiments.
    Other settings remain unchanged.
    Since CUCB performs better than Thompson Sampling and $\epsilon$-Greedy on both datasets, we display only the results from CUCB in later experiments.

    \begin{figure}[tp!]
        \centering
        \subfigure[$\textrm{F}_1$ score w.r.t $\tau$ on the Chinese OEI dataset.]{
            \includegraphics[scale=0.45]{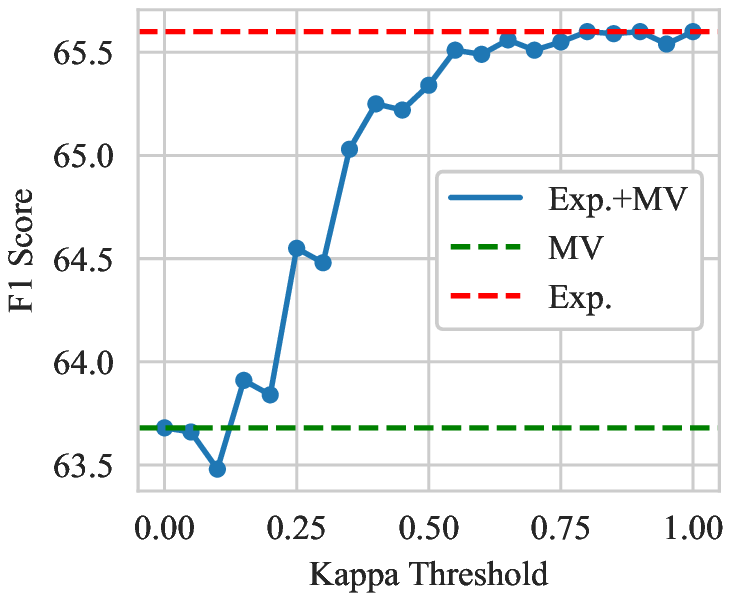}
            \label{fig:oei-f1}
        }
        \subfigure[Expert usage w.r.t $\tau$ on the Chinese OEI dataset.]{
            \includegraphics[scale=0.45]{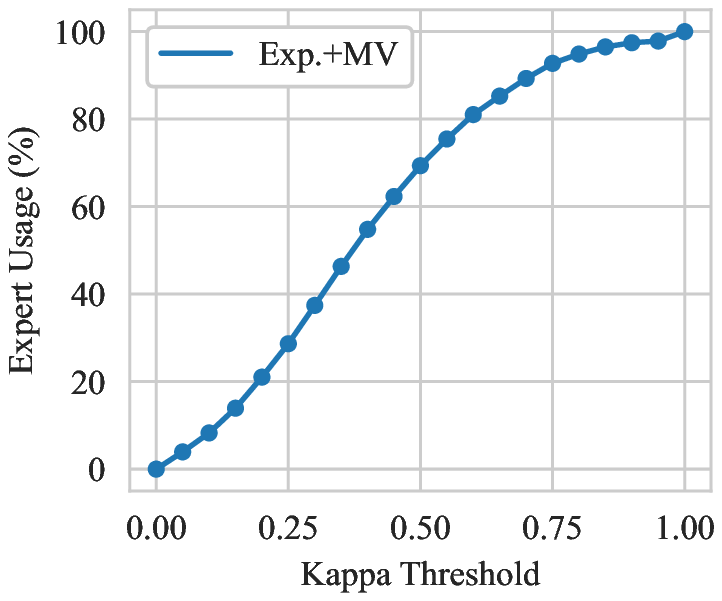}
            \label{fig:oei-expct}
        }
        \subfigure[$\textrm{F}_1$ score w.r.t $\tau$ on the CoNLL 2003 dataset.]{
            \includegraphics[scale=0.45]{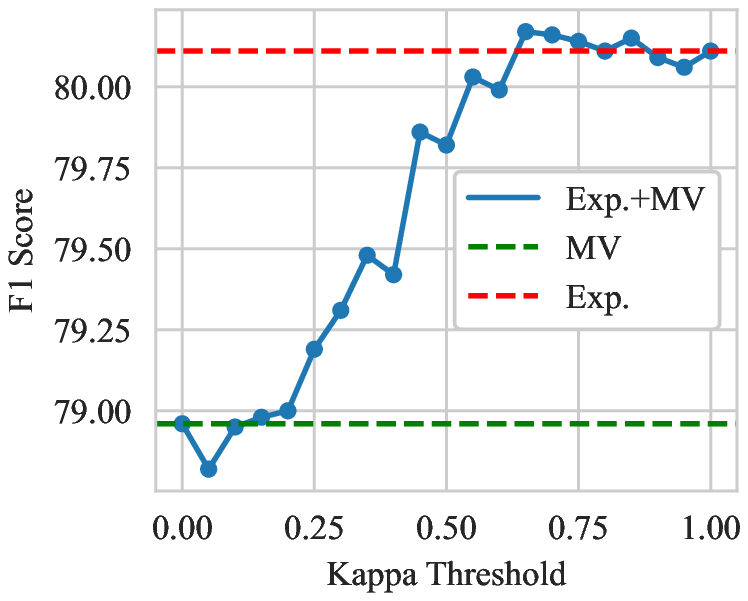}
            \label{fig:conll-f1}
        }
        \subfigure[Expert usage w.r.t $\tau$ on the CoNLL 2003 dataset.]{
            \includegraphics[scale=0.45]{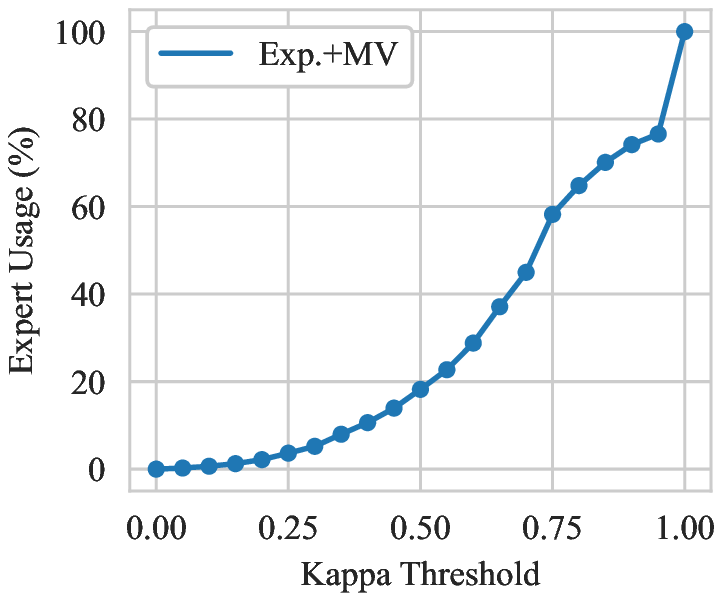}
            \label{fig:conll-expct}
        }
        \caption{$\textrm{F}_1$ scores of the produced annotations and usage of expert for annotation evaluations w.r.t the kappa threshold $\tau$ of the \textbf{Exp.+MV} method on Chinese OEI and CoNLL 2003 datasets.}
        \label{fig:curves}
    \end{figure}


    On the Chinese OEI dataset, as illustrated in Figure~\ref{fig:oei-f1} and~\ref{fig:oei-expct}, $\textrm{F}_1$ increases sharply with $\tau \in [0, 0.4]$.
    When $\tau = 0.4$, \textbf{Exp.+MV} achieves 99.47\% $\textrm{F}_1$ score of \textbf{Exp.}, and saves 47.19\% of the expert cost.
    The $\textrm{F}_1$ score goes up slowly until $\tau$ reaches 0.8.
    When $\tau = 0.8$, the $\textrm{F}_1$ score of \textbf{Exp.+MV} becomes exactly the same as the one of \textbf{Exp.}, and \textbf{Exp.+MV} still saves 6.6\% of the expert cost.


    The results on the CoNLL 2003 dataset are shown in Figure~\ref{fig:conll-f1} and~\ref{fig:conll-expct}.
    Similarly, the $\textrm{F}_1$ score of the produced annotation dataset grows fast as $\tau \in [0, 0.45]$.
    When $\tau = 0.45$, the \textbf{Exp.+MV} method already produce an annotation dataset with its $\textrm{F}_1$ reaching 99.86\% of \textbf{Exp.}.
    At this point, \textbf{Exp.+MV} saves 88.57\% of the expert cost.
    When $\tau = 0.65$, \textbf{Exp.+MV} outperforms \textbf{Exp.} with a 100.04\% $\textrm{F}_1$ score and a 65.97\% reduction in expert usage.

    Our \textbf{CUCB~(Exp.+MV)} worker selection algorithm eliminates the need for expert evaluation on every sentence.
    Instead, we harness crowd intelligence via our kappa-thresholded MV, producing datasets of comparable or even superior quality to those relying solely on expert evaluations.

    \paragraph{Extended $\textrm{F}_1$ Metrics} All of the $\textrm{F}_1$ scores in the previous experiments are span-level proportional scores calculated by the proportion of the overlap referring to the expert annotation~\cite{zhang-etal-2022-identifying}.
    To provide additional comparisons between different methods, we also invoke token-level and span-level exact $\textrm{P}$, $\textrm{R}$, $\textrm{F}_1$ scores as supporting metrics.
    We run the whole process from data augmentation to worker selection with all 3 metrics separately.
    The kappa threshold $\tau$ in \textbf{Exp.+MV} is set to 0.4 on the Chinese OEI dataset and 0.65 on the CoNLL 2003 dataset.
    Detailed scores are listed in Table~\ref{table:conll-results} and~\ref{table:oei-results}.
    The results show that \textbf{Exp.+MV} achieves scores as good as \textbf{Exp.} and much better than \textbf{MV}, which validates previous experiments and shows our worker selection methods are robust to different metrics.

    \paragraph{Feedback Simulator} We also test our worker selection methods with a feedback simulator.
    The simulator generates numerical feedback from $Bernoulli$ distribution in annotation evaluations.
    This is to eliminate the varying level of difficulty in different tasks and evaluate our worker selection algorithms under more stable settings.
    Our algorithm achieves good results on the simulator.
    We put the definitions and results in Appendix~\ref{sec:feedback-simulator}.

    \paragraph{Effect on ML Models} To further show the effect of our worker selection algorithm on the performance of machine learning models,
    we have run experiments with several widely-accepted models and provide the results in Table~\ref{table:model-acc}.
    We observe a consistent increment of F1 score on the ML models, with our bandit-based worker selection algorithm.
    This validates that our worker selection algorithm may help improve the performance of ML models while saving budget on data crowdsourcing.

    \section{Conclusion}\label{sec:conclusion}

    In this study, we introduced a CMAB-based worker selection strategy tailored for span-based sequence labeling tasks, leveraging the span-level $\textrm{F}_1$ with \textbf{Exp.+MV} as a feedback mechanism.
    To address the challenges posed by unbalanced and limited real datasets, we innovated a data augmentation method.
    This technique not only facilitates offline simulation but also mirrors the genuine annotation behaviors of workers closely.

    Our empirical evaluations underscore the efficacy of the proposed method.
    On the Chinese OEI dataset, our approach achieved an impressive 99.47\% $\textrm{F}_1$ score, translating to a substantial 47.19\% reduction in expert costs.
    Similarly, on the CoNLL 2003 dataset, we observed a remarkable 100.04\% $\textrm{F}_1$ score, with savings of up to 65.97\% in expert costs, both benchmarks set against expert-evaluation-only baselines.
    Furthermore, our method demonstrated its robustness with a 94.86\% $\textrm{F}_1$ score and a 65.97\% reduction in expert costs on a data-free simulator.
    Our approach also boosts ML model performance, optimizing both accuracy and cost.

%

    \bibliographystyle{ccl}
    \bibliography{custom}

    \appendix

    \section{Feedback Simulator}
    \label{sec:feedback-simulator}
    The performance of crowd workers can vary across different types of annotation tasks.
    To evaluate the \textbf{Exp.+MV} worker selection method in more stable conditions without task-specific influence, we do not actually annotate the sentences, but directly use a worker's average $\textrm{F}_1$ score to simulate his score on each sentence he annotates.
    The simulated scores are used as the numerical feedback for worker selection.
    Specifically, for each worker $w$, we calculate in advance two average $\textrm{F}_1$ scores for all of their annotations on the original dataset.
    The two $\textrm{F}_1$ scores for each worker are calculated using expert and majority vote (MV) evaluation respectively, denoted as $\Bar{\varphi}_w^{Exp.}$ and $\Bar{\varphi}_w^{MV}$.
    At each time step $t$, for every sentence $s_i$ in the sentence set to be annotated $S_t$, we ask $K$ different workers from the current selected workers $W_t$ to annotate it.
    Then, we use a random value between 0 and 1 as the agreement level $\kappa$.
    If $\kappa$ exceeds the threshold value $\tau$ (set to 0.4 in \textbf{Exp.+MV}), we independently generate feedback for the $K$ workers from a Bernoulli distribution with a probability parameter set to $\Bar{\varphi}_w^{MV}$.
    If not, the feedback is generated from a Bernoulli distribution with a probability parameter set to $\Bar{\varphi}_w^{Exp.}$.
    The span-level average $\textrm{F}_1$ scores of the annotated dataset using different worker selection algorithm are shown in Table~\ref{tab:simulator-results}.
    Our feedback mechanism \textbf{Exp.+MV} for worker selection achieved comparable performance to the expert-only mechanism \textbf{Exp.} (68.29 versus 69.78), while in the same time reduced expert involvement in evaluation by 59.88\% under the dataset-independent conditions.

    \begin{table}[tp]
        \centering
        \begin{tabular}{c|ccc}
            \toprule
            \textbf{Method} & $\textrm{F}_1$ \\
            \midrule
            Oracle          & 74.12          \\
            Random          & 65.12          \\
            Exp.            & \textbf{69.78} \\
            MV              & 66.80          \\
            Exp.+MV         & \textbf{68.29} \\
            \bottomrule
        \end{tabular}
        \caption{The overall span-level proportional $\textrm{F}_1$ scores of all methods with the feedback simulator.}
        \label{tab:simulator-results}
    \end{table}

    \begin{table}[tp!]
        \centering
        \begin{tabular}{l|cc}
            \toprule
            \textbf{Method}   & Original & w/ Our Alg. \\
            \midrule
            LSTM-Crowd-cat~   & 52.66    & 54.27       \\
            Bert-BiLSTM-CRF   & 52.14    & 54.51       \\
            Annotator-Adaptor & 53.86    & 56.16       \\
            \bottomrule
        \end{tabular}
        \caption{Span-level exact $\textrm{F}_1$ scores of widely-accepted deep learning models on the Chinese OEI dataset.
        LSTM-Crowd-cat is from Nguyen et al.~\shortcite{nguyen-etal-2017-aggregating}.
        Bert-BiLSTM-CRF and Annotator-Adaptor are from Zhang et al.~\shortcite{zhang-etal-2022-identifying}.
        We provide results with and without our worker selection algorithm.}
        \label{table:model-acc}
    \end{table}

    \begin{algorithm}[tp]
        \caption{The worker selection algorithm with the Expert+MV metric.}
        \label{alg:cucb}
        \begin{algorithmic}[1]
            \STATE Let each worker $w_j \in W$ annotate a random sentence and initialize variable $\bar{\mu}_j$ with $\textrm{F}_1$ by expert evaluation
            \STATE For each worker $w_j \in W$, initialize $T_j \leftarrow 1$
            \STATE $t \leftarrow \lvert W \rvert$
            \WHILE{unannotated sentences exist}
            \STATE $t \leftarrow t + 1$
            \STATE Select $W_t \subset W$ based on certain criterion~(e.g.,~(\ref{eqn:eg-selection}),~(\ref{eqn:cucb-selection}))
            \STATE Split $W_t$ into several disjoint subsets $\{W_{t1}, \dots, W_{ti}, \dots , W_{tn}\}$, each containing $K$ workers
            \FORALL{$W_{ti}$}
            \STATE Let each $w_j \in W_{ti}$ annotate an sentence $s_i$ and collect the annotations $\boldsymbol{A}_i$
            \IF{$\kappa(\boldsymbol{A}_i) > \tau$}
            \STATE Update $T_j$ and $\bar{\mu}_j$ with $\textrm{F}_1^\textrm{MV}(\boldsymbol{a}_{ij}(t))$
            \ELSE
            \STATE Update $T_i$ and $\bar{\mu}_j$ with $\textrm{F}_1^\textrm{Exp}(\boldsymbol{a}_{ij}(t))$
            \ENDIF
            \ENDFOR
            \ENDWHILE
        \end{algorithmic}
    \end{algorithm}

    \section{Regret Analysis}
    \label{sec:regret-analysis}
    We provide a brief regret analysis of the worker selection framework assuming that we use the $\epsilon$-greedy algorithm and that each worker's reward follows a Bernoulli distribution.

    The main proof follows the proof of Theorem 1 in~\cite{garcelon2022top}. The key contribution here is that we need to specify that the evaluation signal (generated by majority voting) is a generalized linear model of workers' true reward signal (generated by expert/oracle). To this end, we utilize the following form of the Chernoff bound which applies for any random variables with bounded support.
    \begin{Lem}
        \label{chernoff}(Chernoff Bound~\cite{motwani1995randomized})
        Let $X_1, X_2, \cdots, X_N$ be independent random variables such that $x_l \le X_i \le x_h$ for all $i \in \{1,2, \cdots, N\}$. Let $X=\sum_{i=1}^{N}X_i$ and $\mu=\mathbb{E}(X)$. Given any $\delta>0$, we have the following result:
        \begin{equation}
            \label{inequality}
            P(X\le(1-\delta)\mu)\le e^{-\frac{\delta^2 \mu^2}{N(x_h-x_l)^2}}.
        \end{equation}
    \end{Lem}
    For the purpose of our discussion, let $X_i \in \{0,1\}$ be a binary random variable, where $X_i=0$ denotes that worker $i$ provides an incorrect solution, and $X_i=1$ denotes that worker $i$ generates a correct solution. Define $X=\sum_{i\in \mathcal{N}}X_i$.


    We aim to approximate $P_{\rm MV}$, which is the probability that the majority of the $N$ workers provide the correct estimate.
    We apply the Chernoff Bound in Lemma \ref{chernoff} to $P_{\rm MV}$. We can compute
    \begin{equation}
        \mathbb{E}(X)=\bar{p}=\frac{\sum_{i=1}^{N}p_i}{N} .
    \end{equation}
    Based on (\ref{inequality}), we let $\mu=\mathbb{E}(X), \;\delta=\frac{N(\bar{p}-\frac{1}{2})}{\frac{N}{2}+N(\bar{p}-\frac{1}{2})}$, $x_l=0$, $x_h=1$, and get the following result:
    \begin{align}
        &P_{\rm MV}=P\left(X \ge \frac{N}{2}\right)=1- P\left(X \le \frac{N}{2}\right)\notag\\
        &\ge 1-e^{-\frac{\delta^2 \mu^2}{N}}\\
        &=1-e^{-\frac{\frac{N^2(\bar{p}-\frac{1}{2})^2}{[\frac{N}{2}+N(\bar{p}-\frac{1}{2})]^2}[\frac{N}{2}+N(\bar{p}-\frac{1}{2})]^2}{N}}\\
        &=1-e^{-\frac{N^2(\bar{p}-\frac{1}{2})^2}{N}}\\
        &=1-e^{-N\left(\frac{\sum_{i=1}^{N}p_i}{N}-\frac{1}{2}\right)^2}.\label{lower_bound}
    \end{align}

    Through approximating $P_{\rm MV}$ by its lower bound in (\ref{lower_bound}), we can see that the evaluation signal (represented by $P_{\rm MV}$) is an increasing function in each worker's capability $p_i$ and twice-differentiable. That is, $P_{\rm MV}$ is
    a generalized linear function, which satisfies Assumption 3 in~\cite{garcelon2022top}. Therefore, one can follow the proof of Theorem 1 in~\cite{garcelon2022top} that the $\epsilon$-greedy algorithm yields a sub-linear regret with order $\tilde{O}(T^{2/3})$.

    \begin{algorithm}[tp]
        \caption{The annotation selection algorithm.}
        \label{alg:fakeannopick}
        \begin{algorithmic}[1]
            \STATE For each worker $w_j \in W$, maintain (1)a variable $\hat{\varphi}_j$ as the average $\textrm{F}_1$ score of the selected annotations by $w_j$ so far, (2)a set $\boldsymbol{A}^j$ of selected annotations by $w_j$
            \STATE Generate all possible annotations $\boldsymbol{A}_1^p$ on $s_1 \in S$, calculate $\textrm{F}_1^\textrm{Exp}(\boldsymbol{a}_{1k})$ for each $\boldsymbol{a}_{1k} \in \boldsymbol{A}_1^p$
            \STATE For each $w \in W$, initialize $\hat{\varphi}_j$ with the $\textrm{F}_1^\textrm{Exp}(\boldsymbol{a}_{1k})$ closest to $\Bar{\varphi}_j$, and append the $\boldsymbol{a}_{1k}$ to $\boldsymbol{A}^j$
            \FORALL{$s_i \in S \backslash s_1$}
            \STATE Generate all possible annotations $\boldsymbol{A}_i^p$ on $s_i \in S$, calculate $\textrm{F}_1^\textrm{Exp}(\boldsymbol{a}_{ik})$ for each $\boldsymbol{a}_{ik} \in \boldsymbol{A}_i^p$
            \FORALL{$w_j \in W$}
            \IF{$\hat{\varphi}_j > \Bar{\varphi}_j $}
            \STATE Update $\hat{\varphi}_j$ with the maximal $\textrm{F}_1^\textrm{Exp}(\boldsymbol{a}_{ik})$ less than $\Bar{\varphi}_j$, and append $\boldsymbol{a}_{ik}$ to $\boldsymbol{A}^j$
            \ELSE
            \STATE Update $\hat{\varphi}_j$ with the minimal $\textrm{F}_1^\textrm{Exp}(\boldsymbol{a}_{ik})$ greater than $\Bar{\varphi}_j$, and append $\boldsymbol{a}_{ik}$ to $\boldsymbol{A}^j$
            \ENDIF
            \ENDFOR
            \ENDFOR
        \end{algorithmic}
    \end{algorithm}

    \section{Case Study of Annotation Errors}
    \label{sec:case-study}

    Based on our statistical analysis of the Chinese OEI dataset, we find that 74.80\% of annotations have different types of errors.
    And these annotation errors could be decomposed to three basic error types, namely Shifting, Expanding, and Shrinking (SES).
    In our data augmentation algorithm, we reversely used SES modifications and their combinations on the ground truth annotations to generate annotations with varying errors made by crowd workers.
    In this section, we provide a detailed characterization of human-made errors observed on annotated data with real cases to better motivate these modifications.

\begin{figure}[h]
        \centering
        \includegraphics[scale=0.45]{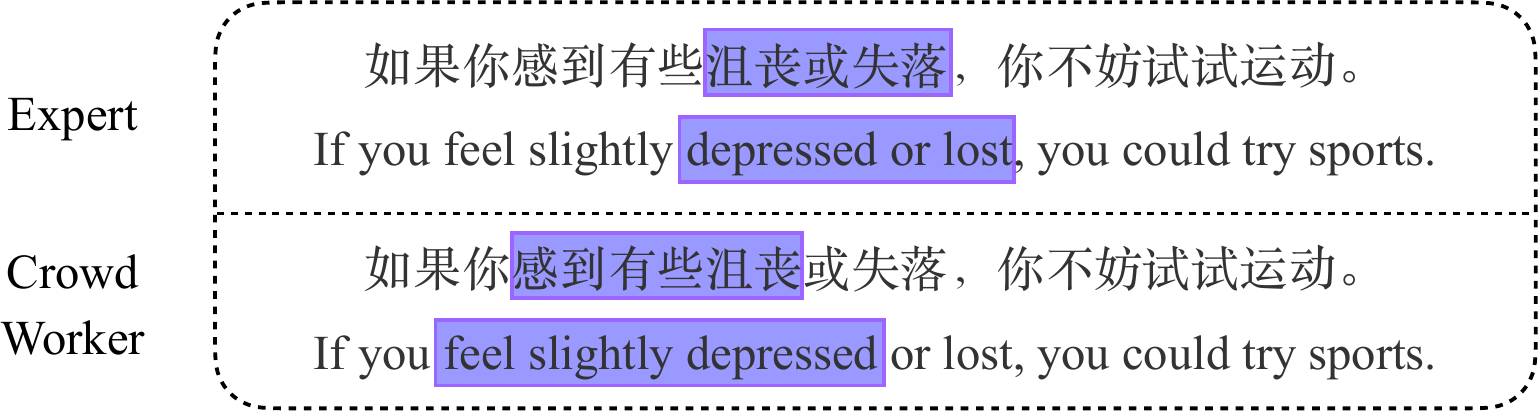}
        \caption{A case in which the crowd worker annotates a span with correct length and polarity but incorrect position.}
        \label{fig:case-shift}
    \end{figure}

    \paragraph{Shifting}
    Some crowd annotation spans are as long as expert ones, but their positions are wrong.
    \textit{Shifting} simulates this type of error.
    As depicted in Figure~\ref{fig:case-shift}, both the expert span and the crowd span are three words long and of negative polarity.
    The difference is that the crowd span is shifted to the left by 2 words compared with the expert span.
    This type of error can be generated with \textit{Shifting} on the expert annotations.

\begin{figure}[h]
        \centering
        \includegraphics[scale=0.45]{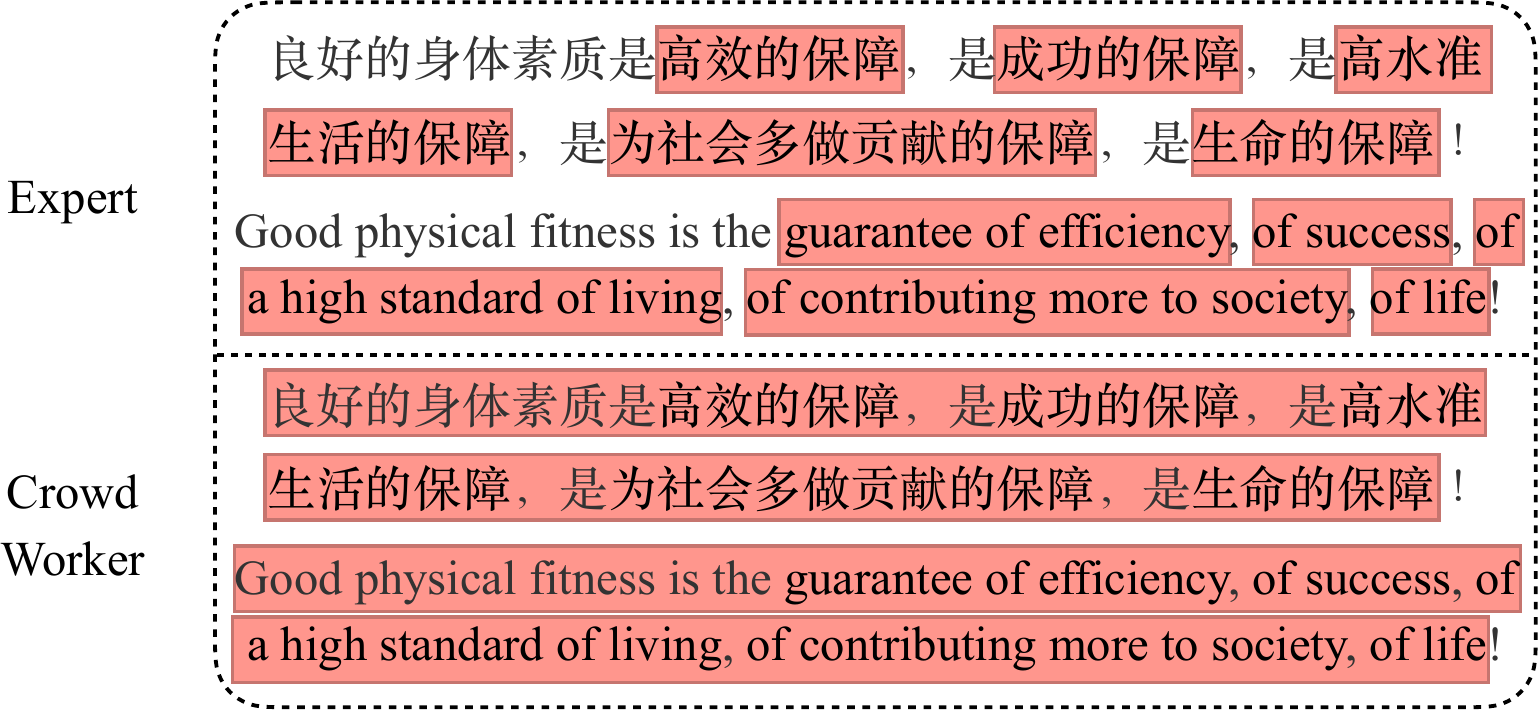}
        \caption{A case in which the crowd worker uses a very long span that covers the whole sentence.}
        \label{fig:case-expand}
    \end{figure}

    \paragraph{Expanding}
    \textit{Expanding} is used to generate longer (than expert span) error spans.
    It might be intuitive that annotators barely make errors such as expanding to a very long span.
    However, in the case illustrated in Figure~\ref{fig:case-expand}, the expert annotates five short spans separated by commas, while the crowd worker uses a very long span that covers the whole sentence, which is obviously not accurate.
    To simulate such human-made errors, we can expand an expert span to cover unnecessary words.
    Statistically, 4.03\% of annotation errors are very long spans with more than 15 Chinese characters.
    So we do not set an upper bound of span length in \textit{Expanding}.

\begin{figure}[h]
        \centering
        \includegraphics[scale=0.45]{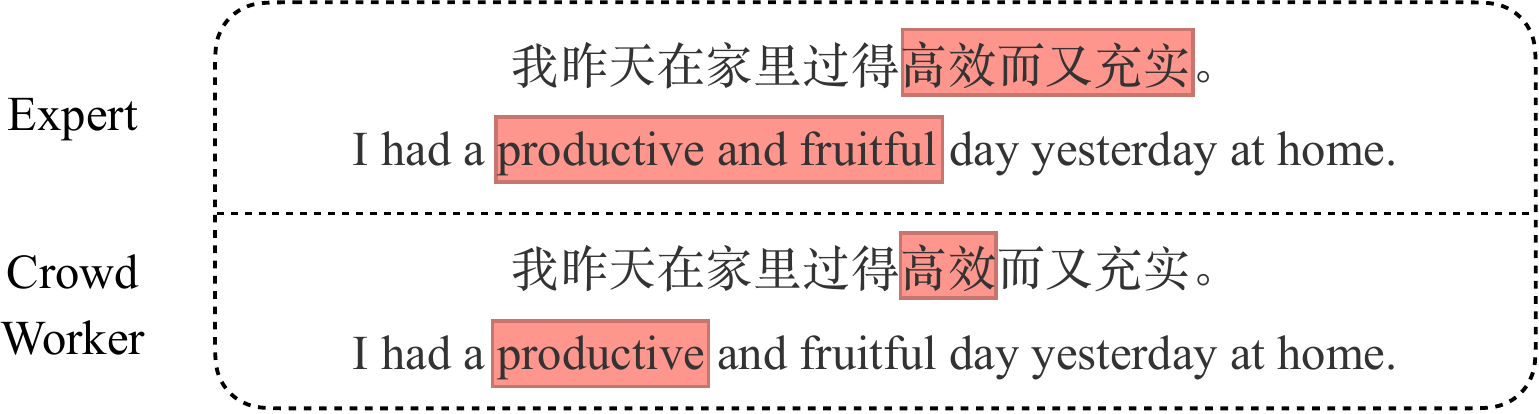}
        \caption{A case in which the crowd worker does not annotate all words with polarity.}
        \label{fig:case-shrink}
    \end{figure}

    \paragraph{Shrinking}
    \textit{Shrinking} is useful since crowd workers often ignore some words when annotating.
    As shown in Figure~\ref{fig:case-shrink}, the crowd worker failed to find all words expressing positive opinions.

    Sometimes crowd workers ignore a whole span in expert annotations.
    This is why we set the lower bound of span length to zero in \textit{Shrinking}, which means we can shrink a span into no span.

    These three types of errors may occur separately or combined in real crowd annotations.
    Such that an error could be both shifted and shrunk.
    This is why we use the combination of these three types of modifications to simulate human-made errors in our data augmentation algorithm.

    \section{A Running Example of Data Augmentation}\label{sec:augmentation-running-example}

    We here provide a running example to illustrate how an annotation for a certain worker on a certain sentence is generated with our proposed augmentation method. Suppose we have an English sentence:

    \begin{quote}
        Although he looked very depressed yesterday, he has already become much more cheerful now.
    \end{quote}

    And an expert annotation:

    \begin{quote}
        Although he looked \textbf{[NEGATIVE: very depressed]} yesterday, he has already become \textbf{[POSITIVE: much more cheerful]} now.
    \end{quote}

    If the crowd worker Sam has an annotation on this sentence in the original dataset, we use it directly in the augmented dataset. Otherwise, we generate an annotation for Sam with our data augmentation method.

    When generating annotation for Sam, we follow the steps below:

    \begin{enumerate}
        \item For each span in the expert annotation, we apply the Shifting, Expanding, and Shrinking (SES) modifications on it. After this step, we have several lists of annotation, each list contain annotations with only one modified span:

        \begin{itemize}
            \item List 1, modifications of the first span, containing \(N_1\) annotations:
            \begin{itemize}
                \item Although he looked \textbf{[NEGATIVE: very depressed]} yesterday, he has already become much more cheerful now. \textit{\# Unmodified, span-level proportional F1 = 1.0}

                \item Although he looked very \textbf{[NEGATIVE: depressed yesterday]}, he has already become much more cheerful now. \textit{\# Shifting, F1 = 0.5}
                \item Although he looked very depressed \textbf{[NEGATIVE: yesterday ,]} he has already become much more cheerful now. \textit{\# Shifting, F1 = 0}
                \item [...] \textit{\# Other Shifting modifications}

                \item Although he \textbf{[NEGATIVE: looked very depressed]} yesterday, he has already become much more cheerful now. \textit{\# Expanding, F1 = 1.0}
                \item [...] \textit{\# Other Expanding modifications}

                \item Although he looked very \textbf{[NEGATIVE: depressed]} yesterday, he has already become much more cheerful now. \textit{\# Shrinking, F1 = 0.5}
                \item [...] \textit{\# Other Shrinking modifications}
            \end{itemize}

            \item List 2, modifications of the second span, containing \(N_2\) annotations:
            \begin{itemize}
                \item Although he looked very depressed yesterday, he has already become \textbf{[POSITIVE: much more cheerful]} now. \textit{\# Unmodified, span-level proportional F1 = 1.0}
                \item Although he looked very depressed yesterday, he has already become much \textbf{[POSITIVE: more cheerful now]}. \textit{\# Shifting, F1 = 0.6667}
                \item Although he looked very depressed yesterday, he has already become much more \textbf{[POSITIVE: cheerful now .]} \textit{\# Shifting, F1 = 0.3334}
                \item [...] \textit{\# Other Shifting modifications}

                \item Although he looked very depressed yesterday, he has already \textbf{[POSITIVE: become much more cheerful]} now. \textit{\# Expanding, F1 = 1.0}
                \item [...] \textit{\# Other Expanding modifications}

                \item Although he looked very depressed yesterday, he has already become much \textbf{[POSITIVE: more cheerful]} now. \textit{\# Shrinking, F1 = 0.6667}
                \item [...] \textit{\# Other Shrinking modifications}
            \end{itemize}
        \end{itemize}

        \item We choose one annotation from each list, and combine them to generate an annotation with 2 spans. This is done for all combinations of the annotations in the two lists. Note that if the two spans overlay with each other, we merge them into one span. After step 2, we have one list of annotations:

        \begin{quote}
            Combined List, containing less than or equal to \(N_1 \times N_2\) annotations:
            \begin{itemize}
                \item Although he looked \textbf{[NEGATIVE: very depressed]} yesterday, he has already become \textbf{[POSITIVE: much more cheerful]} now.  \textit{\# span-level proportional F1 = 1.0}
                \item Although he looked very \textbf{[NEGATIVE: depressed yesterday]} , he has already become \textbf{[POSITIVE: much more cheerful]} now.  \textit{\# span-level proportional F1 = 0.75}
                \item Although he looked very depressed \textbf{[NEGATIVE: yesterday ,]} he has already become \textbf{[POSITIVE: much more cheerful]} now.  \textit{\# span-level proportional F1 = 0.5}
                \item [...] \textit{\# Other combinations with F1 ranging from 0 to 1.0}
            \end{itemize}
        \end{quote}

        \item We choose one annotation from the combined list as Sam's annotation on this sentence, according to the following procedure:
        \begin{enumerate}
            \item Sam has an average F1 score \(F_{\text{ori}} = 0.57\) on the original (real) dataset.
            \item We have already got 10 annotations for Sam in the augmented dataset, which has an average F1 score \(F_{\text{aug\_10}} = 0.54\).
            \item We are choosing annotation on the 11th sentence for Sam.
            \item We firstly select two annotations with the closest F1 scores to \(F_{\text{ori}}\) from the combined list, one higher than \(F_{\text{ori}}\), and one lower than \(F_{\text{ori}}\), as candidate annotations. In this case, the two annotations could have F1 scores of 0.58 and 0.52 respectively.
            \item If \(F_{\text{aug\_10}} > F_{\text{ori}}\), we choose the annotation with the lower F1 score (0.52) as Sam's annotation on this sentence. Otherwise, we choose the annotation with the higher F1 score (0.58). This is to ensure that the average F1 score of Sam's annotations in the whole augmented dataset, \(F_{\text{aug}}\), is as close to \(F_{\text{ori}}\) as possible, which reflects Sam's reliability~(i.e., performance). In this case, we choose the annotation with F1 score of 0.58.
        \end{enumerate}
    \end{enumerate}

    By generating the missing annotations in the original dataset with the method above, we could have an augmented dataset.

    \section{Explanation of Worker Selection on Building New Datasets}\label{sec:new-dataset-explanation}

    When creating new datasets, we expect to have a few (e.g. five) experts and a relatively large group of (e.g. a hundred) crowd workers available for annotation.

    At each time step, we select a group of (e.g. 20) crowd workers, and request them to annotate a few (e.g. 5) sentences, resulting in 4 crowd annotations on each sentence. Now we calculate the agreement of the annotations on each sentence, if the agreement is high (e.g. greater than 0.4), we use the MV aggregation of the crowd annotations as the ground truth, and calculate the F1 scores of each worker's annotation. Otherwise, we ask an expert to give an annotation on the sentence, and calculate the F1 score of each worker on the expert annotation. Note that the expert annotates only when the agreement is low. After this time step, we have crowd annotations on the sentences and their F-scores, which can be used to update the average score of each worker. This procedure is repeated until we have enough annotations on every sentence.

    In other words, the Expert+MV approach works with both crowd workers and experts available (e.g. on an online system) when building datasets. The Expert+MV is an iterative approach in which the expert annotates when needed. And it saves the cost of expert annotations by using the MV aggregation of crowd annotations as the ground truth when possible. Our experiment results show that the Expert+MV approach can save 47.19\% of the cost of expert annotations on the Chinese OEI dataset, and 65.97\% on the CoNLL'03 dataset respectively.

    However, even in the case that no expert is available, which means that Expert+MV falls back to MV, we can still observe that the MV approach outperforms the Random baseline (which is an equivalent of normal crowdsourcing procedure which assigns an equal amount of sentences to each worker randomly) by a large gap. In this case, the MV approach saves 100\% of expert annotation cost, but still produced crowd annotation with good quality. Please refer to Table~\ref{table:conll-results} and Table~\ref{table:oei-results} for more detailed results.

    \begin{table*}[tp]
        \centering
        \begin{tabular}{c|p{1.1cm}<{\centering}p{1.1cm}<{\centering}p{1.1cm}<{\centering}p{1.1cm}<{\centering}|c|p{1.1cm}<{\centering}p{1.1cm}<{\centering}p{1.1cm}<{\centering}p{1.1cm}<{\centering}}
            \toprule
            \bfseries\makecell{Worker \\ID} & \bfseries\makecell{Ori. \\$\textrm{F}_1$} & \bfseries\makecell{Rnd. \\Gen. \\$\textrm{F}_1$}  & \bfseries\makecell{SES \\Only \\$\textrm{F}_1$}  & \bfseries\makecell{SES \\+Alg.\ref{alg:fakeannopick} \\$\textrm{F}_1$} & \bfseries\makecell{Worker \\ID} & \bfseries\makecell{Ori. \\$\textrm{F}_1$} & \bfseries\makecell{Rnd. \\Gen. \\$\textrm{F}_1$} & \bfseries\makecell{SES \\Only \\$\textrm{F}_1$} & \bfseries\makecell{SES \\+Alg.\ref{alg:fakeannopick} \\$\textrm{F}_1$} \\
            \midrule
            25 & 62.90 & 60.07 & 69.59 & 62.89 & 37  & 37.15 & 96.10 & 26.79 & 37.16 \\
            32 & 60.87 & 41.37 & 68.79 & 60.87 & 13 & 36.19 & 31.62 & 25.14 & 36.20 \\
            42 & 53.88 & 4.37 & 66.57 & 53.88 & 20 & 36.11 & 71.44 & 25.02 & 36.12 \\
            5  & 52.07 & 50.74 & 60.76 & 52.06 & 64 & 35.97 & 65.66  & 25.39 & 35.97 \\
            55 & 50.70 & 30.24 & 61.13 & 50.70 & 63 & 35.22 & 75.40  & 24.73 & 35.22 \\
            2 & 50.53 & 91.99  & 60.92 & 50.53 & 6 & 35.15 & 65.74  & 25.00 & 35.16 \\
            52 & 50.08 & 41.93 & 60.91 & 50.08 & 10 & 34.63 & 51.28 & 25.08 & 34.64 \\
            17  & 49.82 & 43.73 & 35.82 & 49.82 & 66 & 33.75 & 60.98 & 24.99 & 33.75 \\
            57 & 49.25 & 13.17 & 35.59 & 49.25 & 53 & 32.90 & 27.51 & 24.78 & 32.89 \\
            11 & 49.04 & 53.71 & 35.19 & 49.03 & 4 & 32.72 & 8.40  & 24.77 & 32.72 \\
            26 & 48.89 & 5.17 & 35.59 & 48.82 & 21 & 32.19 & 73.47  & 24.78 & 32.19 \\
            36 & 48.71 & 15.53 & 35.27 & 48.70 & 62  & 32.16 & 48.71 & 24.89 & 32.16 \\
            46 & 48.67 & 44.84 & 35.19 & 48.67 & 1 & 32.10 & 34.42 & 24.96 & 32.10 \\
            29 & 48.60 & 95.39 & 35.21 & 48.60 & 41 & 31.94 & 77.55 & 24.88 & 31.93 \\
            35 & 47.07 & 23.64 & 35.34 & 47.07 & 51 & 31.78 & 68.07 & 24.85 & 31.78 \\
            49 & 46.80 & 60.30 & 35.27 & 46.80 & 31  & 31.61 & 29.44 & 24.59 & 31.61 \\
            54 & 45.63 & 18.74 & 34.45 & 45.64 & 8 & 31.05 & 28.55 & 24.76 & 31.05 \\
            14 & 45.13 & 60.99  & 34.54 & 45.13 & 67 & 30.91 & 95.51 & 24.22  & 30.91 \\
            43 & 44.93 & 34.91 & 33.72 & 44.93 & 58 & 30.70 & 21.64 & 23.96  & 30.70 \\
            7 & 44.37 & 23.89 & 33.50 & 44.37 & 65 & 30.61 & 4.51 & 24.17  & 30.60 \\
            59 & 44.36 & 72.37 & 33.61 & 44.37 & 38 & 30.47 & 4.82 & 24.11  & 30.47 \\
            23 & 43.38 & 4.85 & 33.58 & 43.38 & 28 & 29.86 & 2.63 & 24.00  & 29.86 \\
            56 & 43.37 & 41.96 & 33.31 & 43.37 & 45 & 29.38 & 36.13 & 24.15  & 29.38 \\
            0 & 41.60 & 66.81 & 28.19 & 41.61 & 30 & 28.70 & 61.16 & 21.88  & 28.71 \\
            18 & 41.40 & 31.53 & 28.56 & 41.40 & 15 & 25.73 & 38.92 & 21.40  & 25.73 \\
            16 & 41.31 & 57.13 & 28.03 & 41.31 & 19 & 24.69 & 4.39 & 21.31  & 24.70 \\
            22 & 41.05 & 85.83 & 28.21 & 41.06 & 44 & 23.42 & 7.15 & 21.08  & 23.42 \\
            47 & 40.78 & 82.33 & 27.91 & 40.78 & 9 & 22.88 & 96.22 & 21.22  & 22.89 \\
            61 & 40.22 & 12.20 & 28.44 & 40.22 & 33 & 22.36 & 29.89 & 19.50  & 22.36 \\
            40 & 40.01 & 84.98 & 28.38 & 40.02 & 39 & 20.69 & 57.73 & 19.26  & 20.69 \\
            50 & 39.35 & 56.04 & 28.64 & 39.35 & 69 & 20.39 & 63.02 & 19.26  & 20.40 \\
            27 & 38.77 & 34.07 & 27.87 & 38.77 & 3 & 17.12 & 28.70 & 18.66  & 17.13 \\
            48 & 38.35 & 23.77 & 27.57 & 38.35 & 24 & 16.96 & 42.73 & 18.68  & 16.98 \\
            34 & 38.29 & 5.69 & 28.08 & 38.30 & 68 & 14.53 & 13.63 & 7.69  & 14.53 \\
            12 & 37.96 & 85.14 & 27.44 & 37.96 & 60 & 13.66 & 22.69 & 8.15  & 13.66 \\
            \bottomrule
        \end{tabular}
        \caption{Comparisons between different data augmentation methods on the span-level exact $\textrm{F}_1$ score of every crowd worker.
        \textbf{Ori.} stands for the original score in real datasets before any augmentation.
        \textbf{Rnd. Gen.} is a naive augmentation method with random generated annotations.
        \textbf{SES Only} indicates the \textit{shifting}, \textit{shrinking}, and \textit{expanding} method we proposed.
        \textbf{SES + Alg.\ref{alg:fakeannopick}} means SES with Algorithm~\ref{alg:fakeannopick} which is our final method.}
        \label{table:worker-scores}
    \end{table*}

\end{document}